\documentclass[11pt,twocolumn,letterpaper]{article}
\usepackage[pagenumbers]{wacv} % To force page numbers, e.g. for an arXiv version

\usepackage{amsmath}
\usepackage{amssymb}
\usepackage{booktabs}
\usepackage{adjustbox}
\usepackage[hidelinks]{hyperref}
\usepackage{balance}

\begin{document}

%%%%%%%%% TITLE - PLEASE UPDATE
\title{Adaptive Compensation for Robotic Joint Failures Using Partially Observable Reinforcement Learning}

\author{Tan-Hanh Pham$^{1, *}$, Godwyll Aikins$^{1}$, Tri Truong$^{2}$, and Kim-Doang Nguyen$^{1, *}$\\
\hphantom{text}\\
$^1$Department of Mechanical and Civil Engineering, Florida Institute of Technology, USA \\
$^2$Department of Fundamentals of Machine Design, \\HCMC University of Technology and Education, Vietnam\\
{\tt\small $^{*}$Corresponding authors: hanhpt.phamtan@gmail.com; knguyen@fit.edu}
}
\maketitle

%%%%%%%%% ABSTRACT
\begin{abstract}
Robotic manipulators are widely used in various industries for complex and repetitive tasks. However, they remain vulnerable to unexpected hardware failures. In this study, we address the challenge of enabling a robotic manipulator to complete tasks despite joint malfunctions. Specifically, we develop a reinforcement learning (RL) framework to adaptively compensate for a non-functional joint during task execution. Our experimental platform is the Franka robot with 7 degrees of freedom (DOFs). We formulate the problem as a partially observable Markov decision process (POMDP), where the robot is trained under various joint failure conditions and tested in both seen and unseen scenarios. We consider scenarios where a joint is permanently broken and where it functions intermittently. Additionally, we demonstrate the effectiveness of our approach by comparing it with traditional inverse kinematics-based control methods. The results show that the RL algorithm enables the robot to successfully complete tasks even with joint failures, achieving a high success rate with an average rate of 93.6\%. This showcases its robustness and adaptability. Our findings highlight the potential of RL to enhance the resilience and reliability of robotic systems, making them better suited for unpredictable environments. All related codes and models are published \href{https://hanhpt23.github.io/franka-IK/}{online}.
\end{abstract}

\def\thefootnote{$\$$}\footnotetext{This work is supported by National Science Foundation (NSF) Grant ${\#}$2138206.}

%%%%%%%%% BODY TEXT
\section{Introduction}
\label{sec:intro}

Robotic manipulators are transforming industries across the board, from manufacturing and logistics to healthcare and agriculture. Their precision, versatility, and ability to handle complex tasks make them indispensable in modern automation. As artificial intelligence and sensing technologies advance, these robots are becoming increasingly adaptable, promising even greater impact in the future. The global market for industrial robots reached 16.5 billion USD in 2020, with manipulators being the most widely adopted type \cite{IFR2021}.

However, like any complex system, robotic manipulators are prone to faults. These can result in performance issues, task failures, or even safety hazards. In critical applications such as medical surgery or space exploration, a malfunction could have catastrophic consequences. For instance, a series of incidents involving collaborative robots in automotive plants led to temporary halts in production and raised concerns about human-robot interaction safety \cite{6630576}. This vulnerability underscores the importance of fault-tolerant control (FTC) in robotic systems. FTC is crucial for ensuring continuous operation despite faults or failures, thereby enhancing the reliability and safety of robotic manipulators. As these machines become more prevalent in high-stakes environments, the need for robust fault tolerance mechanisms becomes increasingly vital. 

FTC approaches can be broadly categorized into traditional model-based methods and emerging learning-based techniques \cite{uno1, uno2}. Model-based FTC relies on mathematical representations of the robotic system to detect and address faults \cite{blanke}. These methods typically employ observers or estimators to monitor performance and identify deviations from expected behavior, signaling potential faults. While effective, they may struggle with highly complex systems due to the need for detailed fault models \cite{ding2}. In contrast, learning-based FTC leverages machine learning algorithms to enhance fault tolerance \cite{intelFTC}. This approach can adapt to new and unforeseen faults by learning from data, potentially offering better scalability for complex systems. By utilizing large datasets and sophisticated algorithms, learning-based methods can potentially overcome some limitations of traditional approaches, especially in handling intricate or unexpected fault scenarios. 

Learning-based FTC methods show significant potential, but their current implementation often involves layering machine learning algorithms onto existing control systems \cite{tl1, tl2, tl3}. These algorithms monitor system performance and adjust control actions based on fault detection and diagnosis. This approach offers advantages in modularity, allowing for easier updates or replacements of individual components without overhauling the entire control system. However, this integration strategy has limitations. Response times may be slower due to the additional layer of communication between modules \cite{sardashti2023learning}. The system also requires extensive tuning and training for various operating conditions, potentially limiting its adaptability. Moreover, its effectiveness can be compromised if the underlying mathematical model of the robotic system is inaccurate.

To address these challenges, we propose an innovative end-to-end learning-based framework for fault-tolerant control of robotic manipulators. This approach harnesses the power of deep reinforcement learning (DRL) to create a unified, adaptive, and efficient control system capable of dynamically handling faults. Our proposed system learns to manage faults directly from raw sensory inputs, eliminating the need for separate fault detection, diagnosis, and control modules. 

Importantly, we frame this problem as a partially observable Markov decision process (POMDP). In a POMDP, the agent does not have full information about the state of the environment. This partial observability is particularly relevant in our scenario, where joint malfunctions may occur without explicit notification. The robot must infer the state of its joints from its observations and actions, making decisions under uncertainty.

This integration offers several advantages:
\begin{enumerate}
    \item Unified approach: By combining fault detection, diagnosis, and control into a single system, we potentially reduce complexity and improve response times.
    \item Adaptability: The DRL agent can learn to handle a wide range of faults, including those not explicitly modeled during training, enhancing the system's robustness.
    \item Efficiency: Direct processing of raw sensory data eliminates the need for complex feature engineering or intermediate representations.
    \item Scalability: As the complexity of the robotic system increases, the DRL approach can potentially scale more effectively than traditional methods, given sufficient training data and computational resources.
    \item Continuous learning: The system is designed to update its policies in real time, allowing for ongoing adaptation to new fault scenarios or changing operating conditions.
\end{enumerate}

Our novel framework aims to push the boundaries of fault-tolerant control in robotics, potentially offering a more robust and flexible solution for increasingly complex robotic systems. The rest of the paper is organized as follows. Section \ref{sec:literature} reviews the related work on fault-tolerant control strategies and reinforcement learning in robotic systems. Section \ref{sec:method} presents the details of our proposed methodology, including the problem formulation, the DRL algorithm for fault-tolerant control, and the simulation setup. Compared with the traditional method, section \ref{sec:kinematics} illustrates the failure of inverse kinematic control when one of the joints of a robot is broken. Section \ref{sec:results} described the experimental results obtained from various fault scenarios, demonstrating the effectiveness of our approach. Finally, Section \ref{sec.6conclusion} provides concluding remarks and discusses potential future work.

\section{Literature Review}
\label{sec:literature}

The increasing complexity and autonomy of robotic systems necessitate robust FTC strategies to ensure reliable operation, even in the presence of faults  \cite{blanke}. FTC encompasses strategies and algorithms that enable robotic systems to adapt to faults and continue operation, potentially with degraded performance, rather than experiencing catastrophic failure \cite{Chen1999}. Faults can arise from various sources within a robotic system. Actuator faults, including partial or complete loss of effectiveness, stuck actuators, and total actuator failures, can significantly impair a robot's ability to execute desired movements and interact with its environment \cite{Yao}. Sensor faults, ranging from bias, noise, and drift to complete sensor failures, can compromise a robot's perception of its surroundings, leading to inaccurate localization and decision-making \cite{duan}. Structural faults, such as physical damage or degradation to robot components, can induce changes in dynamic behavior, affecting stability and control performance \cite{saim}. The core objective of FTC is to detect, isolate, and accommodate these faults to maintain overall system stability, performance, and safety. Broadly, FTC approaches are classified into two main categories. Passive FTC methods leverage robust control techniques to design controllers inherently tolerant to a pre-defined range of faults \cite{passive}. While not requiring explicit fault detection and isolation (FDI), their fault tolerance capacity is often limited. Active FTC methods, on the other hand, hinge on real-time FDI to accurately identify and isolate faults \cite{active}. Based on the detected fault information, the controller is reconfigured or adapted online to preserve stability and achieve the desired performance under the given fault conditions.

FTC strategies have been successfully applied to a wide range of robotic systems, demonstrating their versatility in mitigating the impact of various fault types. \cite{sihao1} proposed an innovative incremental nonlinear FTC method for quadcopters experiencing a catastrophic failure: the complete loss of two opposing rotors. This critical scenario typically renders such aerial vehicles inoperable. However, by implementing this advanced FTC strategy, the quadcopter could maintain stable flight and control, albeit with reduced maneuverability, preventing otherwise catastrophic failure.

\cite{ali} proposed an innovative FTC approach that simultaneously addressed both actuator and sensor faults. Their method employed nonlinear backstepping control coupled with friction compensation, enhancing the manipulator's ability to maintain precise movements and positioning even when faced with sensor inaccuracies or actuator inefficiencies. 
Researchers tackled the complex challenge of controlling uncrewed underwater vehicles in the presence of multiple system uncertainties and disturbances. They developed a novel sensor active FTC scheme that achieved trajectory tracking without relying on linear and angular velocity measurements \cite{uuv1}.

Traditional fault-tolerant control systems have several limitations, including their reliance on accurate system models, difficulty in handling complex nonlinear systems, and limited adaptability to unforeseen faults or changing conditions. These systems often struggle with uncertainty and may fail when encountering scenarios not explicitly accounted for in their design \cite{FTC1}. In contrast, learning-based methods offer a more robust and flexible approach to fault-tolerant control. By leveraging data-driven techniques, learning-based FTC can adapt to new situations, learn from experience, and handle complex, nonlinear systems more effectively \cite{survey_ftc}. 

Machine learning (ML) has significantly impacted various areas, greatly improving efficiency and capabilities. In healthcare, ML has enhanced diagnostic accuracy and treatment personalization by analyzing vast patient data and medical images \cite{med3, med1, hanh_med}. In finance, it aids fraud detection and risk assessment, leading to more secure transactions and investment decisions \cite{fraud1, fraud2}. In transportation, machine learning enables autonomous vehicles to navigate complex environments and optimize traffic flow for improved efficiency \cite{car0, car1, car2}. Additionally, machine learning has revolutionized agriculture, enabling precision farming techniques that optimize resource allocation and crop yield prediction \cite{pham2024enhanced, agri1, pham2024soil, hanh_agri}.  Overall, machine learning's ability to learn from data and make intelligent predictions has greatly enhanced fault-tolerant control strategies for robotic systems, offering improved accuracy and adaptability in fault detection, isolation, and accommodation. These methods can potentially identify and respond to a wider range of faults, including those not anticipated during the design phase, and can continuously improve their performance over time \cite{intelFTC}. Additionally, machine learning approaches can better handle the high dimensionality and uncertainty inherent in many modern control systems, making them particularly well-suited for applications in areas such as autonomous vehicles, robotics, and advanced manufacturing \cite{intelFTC}.

Researchers in \cite{ESKI2011115}, demonstrated the use of radial basis neural networks for detecting faults in robotic manipulators, achieving high accuracy by training the network on data from normal and faulty operations. In their work, \cite{gen13} proposed a fault identification method that utilizes multiple source domains to enhance diagnostic accuracy in real-world scenarios. The method effectively learns and transfers generalized diagnostic knowledge from these diverse sources, improving the model's ability to identify faults in new, unseen scenarios.

Supervised and unsupervised ML methods have primarily been used for fault detection, diagnosis, and isolation in robotics. Reinforcement learning (RL) is used for developing adaptive control policies that can respond to faults in real time. However, RL has emerged as a powerful tool for developing adaptive control policies that can respond to faults in real time. RL involves training an agent to make decisions by rewarding desired behaviors and penalizing undesired ones, making it particularly suitable for fault-tolerant control. Researchers compared an RL-based fault-tolerant controller with a model predictive controller (MPC) on a C-130 aircraft fuel tank model. They aimed to test the controllers' adaptability to evolving system changes during operation. Their experiments revealed that the RL-based controller performed more robustly than the MPC under various challenging conditions, including faults, partially observable system models, and fluctuating sensor noise levels \cite{ahmed2018comparison}.

Recent research has demonstrated the effectiveness of RL in various robotic applications. For instance, in \cite{crl}, researchers proposed an adaptive curriculum RL algorithm with dynamics randomization to train a quadruped robot to adapt to random actuator failures. Similarly, \cite{zhu} presented a model-free adaptive fault-tolerant control algorithm based on RL for the multi-joint Baxter robot. Their approach uses parameter estimation and neural networks to identify and compensate for actuator faults and spring interference, thereby improving the robot's tracking performance.
RL has also shown promise in optimizing control parameters in the presence of faults. In \cite{dc_rl}, researchers developed an innovative method to optimize proportional-integral control coefficients specifically for motor position control. This RL-based approach demonstrated superior performance, computational efficiency, and user-friendly implementation compared to the traditional Ziegler-Nichols method, making it accessible even to non-experts.
Expanding the application of RL to aerial robotics, researchers in \cite{aikins} developed a fault-tolerant control system for UAV landing on a moving target. Their approach, which combines robust policy optimization and long-short term memory neural networks, effectively handles sensor failures and noise during the critical landing phase.

While examples highlight the versatility and effectiveness of machine learning in addressing various aspects of fault-tolerant control in robotics, most learning-based FTC approaches fuse machine learning algorithms with existing model-based control systems. This integration, although beneficial in many aspects, still presents several challenges and limitations. The hybrid approach relies on the accuracy of the underlying model-based control system. If the initial model is flawed or oversimplified, the machine learning component may struggle to compensate fully for these inaccuracies. 

This study addresses several key limitations and gaps in previous fault-tolerant control research for robotic manipulators. While prior work has explored model-based approaches and machine learning techniques layered on top of existing control systems, this study proposes a novel end-to-end reinforcement learning framework that directly learns adaptive control policies from raw sensory inputs. Unlike traditional methods that rely on accurate system models or separate fault detection and diagnosis modules, our approach unifies fault handling and control into a single adaptive system. By framing the problem as a POMDP and leveraging the state-of-the-art RL algorithm, we enable the robot to dynamically compensate for both permanent and intermittent joint failures without requiring explicit fault models. Furthermore, our extensive evaluation in both seen and unseen failure scenarios, including challenging cases of partial joint functionality, demonstrates the robustness and generalizability of the proposed approach. By comparing against traditional inverse kinematics methods, we highlight the superior adaptability of our learning-based solution.

\section{Methodology}
\label{sec:method}

\subsection{Problem Formulation}
\label{problem_formulation}
Markov decision processes (MDPs) are mathematical frameworks used to make optimal decision in situations where outcomes are partly random and partly under the control of a decision-maker \cite{sutton2018reinforcement}. An MDP is defined by a set of states, a set of actions, a transition function that determines the probability of moving from one state to another given an action, and a reward function that assigns a numerical reward for each action taken in a given state. The goal in an MDP is to find a policy - a mapping from states to actions - that maximizes the expected sum of rewards over time.

In many real-world scenarios, however, the agent does not have full observability of the environment’s state. This leads to the framework of partially observable Markov decision processes \cite{albrecht2024multi}. A POMDP extends an MDP by incorporating a set of observations and an observation function that provides a probability distribution over possible observations given the actual state and the action taken. The agent must then maintain a belief state, a probability distribution over all possible states, based on the history of actions and observations. This belief state serves as a sufficient statistic for making decisions in a POMDP.

In this study, we frame the problem of a robotic manipulator completing tasks with joint malfunctions as a POMDP. The formulation allows us to account for uncertainties in joint functionality and train the robot to adapt to varying conditions. Our experimental platform utilizes the Isaac Lab environment, which provides a high-fidelity simulation for robotic manipulation tasks \mbox{\cite{mittal2023orbit}}. The platform allows for precise control and measurement of the robot's movements, as well as the ability to simulate various joint failure scenarios. The robot used in our experiments is Franka Emika's Panda, a 7-DOF robotic manipulator known for its precision and dexterity. The Franka robot modeled in Isaac Lab is accurately simulated to match the physical characteristics and kinematic properties of the real robot. The task environment is set up to simulate a typical industrial workspace where the Franka robot is required to open a drawer. The following subsections describe details of the reinforcement learning framework and reward functions.

\subsubsection{Observation Space}
The observation $s_t$ at time $t$ comprises the current joint angles and velocities of both the robot and the drawer, as well as the distance between the robot's gripper and the drawer. Formally, the observation can be represented as:
\begin{equation}
s_t = [\theta_1, \theta_2, \ldots, \theta_9, \dot{\theta}_1, \dot{\theta}_2, \ldots, \dot{\theta}_9, \delta_d, \dot{\delta}_d, \Bar{x}_b, \Bar{y}_b, \Bar{z}_b], 
\end{equation}
where $ \theta_i $ and $ \dot{\theta}_i $ represent the angle and velocity of the $ i $-th joint of the robot, respectively. $\delta_d$ and $\dot{\delta}_d$ are the position and velocity of the drawer. $\Bar{x}_b$, $\Bar{y}_b$, and $\Bar{z}_b$ are the distance between the robot gripper and the drawer in $x$, $y$, and $z$ direction, respectively.

\subsubsection{Action Space}
The action $ a_t $ at time $ t $ consists of the control inputs to the robot's joints and the end-effector positions. The action space is defined as:
\begin{equation}
a_t = [\theta_1, \theta_2, \ldots, \theta_9].
\end{equation}
% where $\theta_i$ is the control input for the $ i $-th joint. 
It is important to note that we cannot apply any arbitrary angle and velocity to the joints. Therefore, we need to trim the angles to their respective upper and lower limits for each joint. To ensure the angles remain within their allowed range, we apply the following constraints:
\begin{equation}
\theta_i^{\text{min}} \leq \theta_i \leq \theta_i^{\text{max}} \quad \text{for} \quad i = 1, 2, \ldots, 9,
\end{equation}
where, $\theta_i^{\text{min}}$ is the minimum allowed angle for the $i$-th joint. $\theta_i^{\text{max}}$ is the maximum allowed angle for the $i$-th joint. By applying these constraints, we ensure that the control inputs are feasible and within the physical limitations of the robot's joints.

\subsubsection{Reward Function}
\label{reward_function}
\textbf{The distance reward} is designed to encourage the robot to minimize the distance between the robot's gripper and the drawer's handle. It is calculated as follows:

\begin{equation}
d = \| \mathbf{p}_{\text{gripper}} - \mathbf{p}_{\text{drawer}} \|_2,
\end{equation}

\begin{equation}
r_{\text{dist}} = \frac{1}{1 + d^2},
\end{equation}
where, $\mathbf{p}_{\text{gripper}}$ is the position of the robot's grasp, $\mathbf{p}_{\text{drawer}}$ is the position of the drawer's handle, and $\| \cdot \|_2$ denotes the Euclidean distance. %Additionally, if $d \leq 0.02$:

\textbf{The rotation reward} is designed to align the robot's gripper orientation with the drawer's handle. It is calculated using the dot products of the forward and up axes:

\begin{equation}
\text{dot}_1 = (\mathbf{a}_1 \cdot \mathbf{a}_2)
\end{equation}

\begin{equation}
\text{dot}_2 = (\mathbf{a}_3 \cdot \mathbf{a}_4),
\end{equation}
where $\mathbf{a}_1$ and $\mathbf{a}_2$ are the gripper forward axis of the Franka and the inward axis of the drawer, respectively. $\mathbf{a}_3$ and $\mathbf{a}_4$ are the gripper upper axis and the drawer, respectively. Eventually, the reward for matching the orientation of the hand to the drawer is expressed as follows:
\begin{equation}
r_{\text{rot}} = 0.5 \left( \text{sign}(\text{dot}_1) \times \text{dot}_1^2 + \text{sign}(\text{dot}_2) \times \text{dot}_2^2 \right).
\end{equation}

\textbf{The around handle reward} ensures the robot's fingers are positioned appropriately around the drawer's handle. If the left finger of the Franka is above the drawer handle and the right below the drawer, we bonus 0.5 for the reward function.

\begin{equation}
r_{\text{handle}} = 
\begin{cases} 
0.5 & \text{if } \mathbf{p}_{\text{left\_finger}, z} > \mathbf{p}_{\text{drawer}, z} \\
    & \text{and } \mathbf{p}_{\text{right\_finger}, z} < \mathbf{p}_{\text{drawer}, z}, \\
0 & \text{otherwise},
\end{cases}
\end{equation}

where, $\mathbf{p}_{\text{left\_finger}, z}$ and $\mathbf{p}_{\text{right\_finger}, z}$ are the z-coordinates of the left and right fingers, respectively.
$\mathbf{p}_{\text{drawer}, z}$ is the z-coordinate of the drawer's grasp position.

\textbf{The open reward} is designed to encourage the robot to open the drawer, which means how far the cabinet has been opened out.

\begin{equation}
r_{\text{open}} = \text{pos}_{\text{drawer\_top}} \times r_{\text{handle}} + \text{pos}_{\text{drawer\_top}},
\end{equation}
where, $\text{pos}_{\text{drawer\_top}}$ is the position of the drawer, which we want the robot to open.

\textbf{The overall reward function} is a combination of the above components, scaled by their respective factors:

\begin{equation}
\adjustbox{scale=0.85}{$
r = w_{\text{dist}} \cdot r_{\text{dist}} + w_{\text{rot}} \cdot r_{\text{rot}} + w_{\text{handle}} \cdot r_{\text{handle}} + w_{\text{open}} \cdot r_{\text{open}},
$}
\end{equation}
where $w_{\text{dist}}, w_{\text{rot}}, w_{\text{handle}}, w_{\text{open}}$ are the scaling factors for distance reward, rotation reward, around handle reward, and open reward, respectively.

This reward structure ensures that the robot is incentivized to minimize the distance to the drawer handle, align its gripper orientation correctly, position its fingers around the handle, and ultimately open the drawer.

\subsection{Reinforcement Learning Framework}
\subsubsection{PPO Algorithm}

In this study, we leverage the Proximal Policy Optimization (PPO) algorithm to train the Franka robot for robust task completion despite joint malfunctions. PPO is an on-policy reinforcement learning algorithm that balances ease of implementation with strong empirical performance. It achieves stable learning by optimizing a clipped surrogate objective, which prevents large, potentially destabilizing updates to the policy.

\begin{figure*}[ht]
    \centering    
    \includegraphics[width=0.75\textwidth]{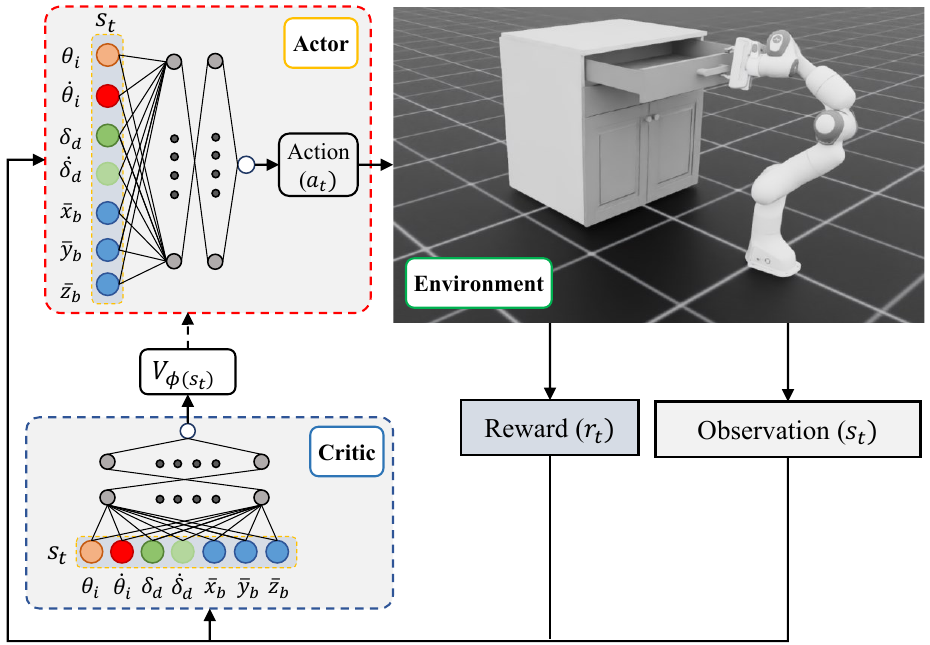}
    \caption{Our framework addresses the issue of joint malfunction in robot manipulators using the PPO algorithm. The Actor takes in the observation from the environment and outputs action for joints. The critic estimates the value of a state, helping the actor learn by providing feedback on the quality of its actions.}
    \label{fig:drl}
\end{figure*}

For PPO, we employed two neural networks: actor and critic networks as illustrated in Figure \mbox{\ref{fig:drl}}. The policy network outputs the robot's action ($a_t$), which represents the angles for each joint ($\theta_i$) of the robot given its current state ($s_t$). Meanwhile, the value network estimates the expected return (value) from a given state. Both networks share an identical architecture comprising three fully connected layers. The first two layers output dimensions of 512 and 256, respectively, followed by ReLU activations. The final layer's output is either the action or the expected return, depending on whether it is the actor or critic network, and it uses a Tanh activation function \mbox{\cite{dubey2022activation}}.

\subsubsection{Agent Training}

The agent's training process involves several key steps, each critical for optimizing the policy and value networks. The workflow for training the agent using the PPO algorithm is as follows:

\textbf{Initialization}: 
Initialize the policy network $\pi_{\vartheta}$ with parameters $\vartheta$. Initialize the value network $V_{\phi}$ with parameters $\phi$. Set the initial parameters for the learning rate, discount factor $\gamma$, and clipping parameter $\epsilon$.

\textbf{Collecting Trajectories}: 
Interact with the environment to collect trajectories of states \(s_t\), actions $a_t$, and rewards \(r_t\). The interaction involves executing actions sampled from the policy network and observing the resulting states and rewards.

\textbf{Computing Returns and Advantages}: 
Calculate the discounted returns \(R_t\) from each time step \(t\):
\begin{equation}
R_t = \sum_{k=0}^{T-t} \gamma^k r_{t+k},
\end{equation}
where \(R_t\) is the return at time step \(t\), \(\gamma\) is the discount factor, and \(r_{t+k}\) is the reward at time step \(t+k\).

Compute the advantage estimates \(A_t\) using the Generalized Advantage Estimation (GAE):
\begin{equation}
A_t = \delta_t + (\gamma \lambda) \delta_{t+1} + \ldots + (\gamma \lambda)^{T-t+1} \delta_T,
\end{equation}
where
\begin{equation}
\delta_t = r_t + \gamma V_{\phi}(s_{t+1}) - V_{\phi}(s_t).
\end{equation}
Here, \(A_t\) is the advantage estimate at time step \(t\), \(\lambda\) is the GAE smoothing parameter, \(\delta_t\) is the temporal difference error, \(V_{\phi}(s_t)\) is the value estimate at state \(s_t\), and \(r_t\) is the reward at time step \(t\).

To calculate the objective loss function of the policy, first of all, we need to calculate the ratio of action probabilities $(r_t(\vartheta)$  between the new and old policies:
\begin{equation}
r_t(\vartheta) = \frac{\pi_{\vartheta}(a_t|s_t)}{\pi_{\vartheta_{\text{old}}}(a_t|s_t)}
\end{equation}
where \(\pi_{\vartheta}(a_t|s_t)\) is the probability of action \(a_t\) given state \(s_t\) under the new policy, and \(\pi_{\vartheta_{\text{old}}}(a_t|s_t)\) is the probability under the old policy. The clipped surrogate objective \(L^{CLIP}(\vartheta)\) is:
\begin{equation}
\adjustbox{scale=0.85}{$
L^{CLIP}(\vartheta) = \mathbb{E}_t \left[ \min \left( r_t(\vartheta) A_t, \text{clip}(r_t(\vartheta), 1 - \epsilon, 1 + \epsilon) A_t \right) \right]
$}
\end{equation}
where \(\mathbb{E}_t\) denotes the expectation over time steps \(t\), and \(\text{clip}(r_t(\vartheta), 1 - \epsilon, 1 + \epsilon)\) restricts \(r_t(\vartheta)\) to the range \([1 - \epsilon, 1 + \epsilon]\).

The total loss for PPO includes both the policy loss (clipped objective) and the squared-error value function loss, as well as an entropy bonus to encourage exploration. The squared-error value function loss is defined as 
\begin{equation}
    L^{value}(\phi) = \mathbb{E}_t \left[ (V_\phi(s_t) - R_t)^2 \right],
\end{equation}

\begin{equation}
\adjustbox{scale=0.85}{$
    \mathcal{L}(\vartheta) = \mathcal{L}^{\text{CLIP}}(\vartheta) - c_1 L^{value}(\phi) + c_2 \mathbb{E}_t \left[ \text{Entropy}(\pi_\vartheta(s_t)) \right].
    $}
\end{equation}

Here, $c_1$ and $c_2$ are coefficients that balance the value loss and entropy bonus. 

After calculating the loss function, we update the policy network parameters by computing the gradients of the total loss $\mathcal{L}(\vartheta)$. Similar to the policy network, we also update the parameters in the value network based on the value loss $L^{value}(\phi)$. 
\begin{equation}
    \vartheta \leftarrow \vartheta - \alpha_\vartheta \nabla_\vartheta \mathcal{L}(\vartheta) 
\end{equation}
\begin{equation}
    \phi \leftarrow \phi - \alpha_\phi \nabla_\phi \mathcal{L}^{value}(\phi). 
\end{equation}
Repeat the update process for a fixed number of epochs or until convergence.

\subsection{Simulation and Experiment Setup}
\label{sec:simulation}

We use the Isaac Lab as the simulation environment for robotic manipulation tasks. By leveraging Isaac Lab, the entire simulation and training process are implemented. The PPO algorithm is implemented by using Pytorch, which is a deep-learning framework. The experiments are conducted on a high-performance computing system with an NVIDIA RTX 2070 Super to accelerate training. This simulation setup allows for thorough testing and validation of our RL framework, demonstrating its potential to enhance the resilience and reliability of robotic systems in the face of unexpected hardware failures.

To evaluate the robustness of our RL framework, we simulate two most realistic types of joint malfunctions:
\begin{itemize}
    \item \textbf{Permanently broken joint}: One of the robot's joints is completely non-functional throughout the task execution.
    \item \textbf{Intermittently functioning joint}: One of the robot's joints operates intermittently, randomly switching between functional and non-functional states.
\end{itemize}

The trained policy is evaluated under both seen and unseen joint failure scenarios to assess its robustness and adaptability. In addition to the intermittent function scenario, we consider two additional test cases for a faulty-functioning joint. In the first test case, the joint is non-functional during the first half of the testing period. In the second test case, the joint is non-functional during the second half of the testing period.

The primary metrics for evaluation include:
\begin{itemize}
    \item Success rate of completing the task (opening the drawer).
    \item Time taken to complete the task.
\end{itemize}
These metrics provide a comprehensive evaluation of the robot's performance under different joint malfunction scenarios. In addition, the performance is compared with a traditional inverse kinematics-based control method to demonstrate the effectiveness of our RL approach.

\section{Inverse Kinematics}
\label{sec:kinematics}
In this section, we describe the kinematic modeling and analysis of the Franka Emika Panda robot used in our experiments. The Denavit-Hartenberg (DH) convention is employed to derive the kinematic equations and inverse kinematics solution for the robotic manipulator. We demonstrate the impact of a joint malfunction by fixing one of the robot's joints and evaluating its ability to perform the task of reaching the desired trajectory as well as opening a drawer.

\subsection{Denavit-Hartenberg Parameters and Inverse Kinematics Solving}

The Franka Emika Panda robot is a 7-DOF manipulator. Each joint is associated with a DH parameter set $(\theta, d, a, \alpha)$, which defines the transformations between consecutive links. The DH parameters for Franka Emika's Panda are summarized in Table~\ref{tab:dh_parameters}.

\begin{table}[ht]
\centering
\caption{DH Parameters for the Franka Emika Panda.}
\label{tab:dh_parameters}
\begin{tabular}{c|c|c|c|c}
\toprule
\textbf{Joint $i$} & $\boldsymbol{d_i}$ (m) & $\boldsymbol{a_i}$ (m) & $\boldsymbol{\alpha_i}$ (rad) & $\boldsymbol{\theta_i}$ (rad) \\
\midrule
1 & 0.333 & 0 & 0 & 1.157 \\
2 & 0 & 0 & $-\pi/2$ & -1.066 \\
3 & 0.316 & 0 & $\pi/2$ & -0.155 \\
4 & 0 & 0.0825 & $\pi/2$ & -2.239 \\
5 & 0.384 & -0.0825 & $-\pi/2$ & -1.841 \\
6 & 0 & 0 & $\pi/2$ & 1.003 \\
7 & 0 & 0.088 & 0 & 0.469 \\
\bottomrule
\end{tabular}
\end{table}

\textbf{The forward kinematics} of the manipulator is obtained by multiplying the homogeneous transformation matrices of each link, derived from the DH parameters. The transformation matrix $A_i$ for the $i$-th joint is given by:
\begin{equation}
\adjustbox{scale=0.8}{$
A_i = \begin{bmatrix}
\cos\theta_i & -\sin\theta_i\cos\alpha_i & \sin\theta_i\sin\alpha_i & a_i\cos\theta_i \\
\sin\theta_i & \cos\theta_i\cos\alpha_i & -\cos\theta_i\sin\alpha_i & a_i\sin\theta_i \\
0 & \sin\alpha_i & \cos\alpha_i & d_i \\
0 & 0 & 0 & 1 \\
\end{bmatrix}.
$}
\end{equation}
The overall transformation from the base frame to the end-effector frame is:

\begin{equation}
T_0^7 = A_1 A_2 A_3 A_4 A_5 A_6 A_7.
\end{equation}

\textbf{The inverse kinematics} involves finding the joint angles $\theta_i$ given the desired position and orientation of the end-effector. This is typically a more complex problem than forward kinematics due to nonlinearity and requires iterative numerical methods or analytical solutions. For the Franka robot, we use an iterative approach based on the Jacobian pseudo-inverse method\cite{dulkeba2013comparison, whitney1969resolved}. The goal is to minimize the error between the current end-effector position/orientation and the desired position/orientation. The update rule for the joint angles is given by:

\begin{equation}
\Delta \theta = J^+ (x_d - x),
\end{equation}
where $\Delta\theta$ is the change in joint angles, $J^+$ is the pseudoinverse of the Jacobian matrix $J$, $x_d$ is the desired end-effector position/orientation, and $x$ is the current end-effector position/orientation.

\subsection{Experiment Setup and Joint Malfunction}
\label{kinematic_setup}
In comparison with the DRL algorithm, we model the problem that the end-effector of the robot goes from its initial position to the drawer, grasping it, and pulling it open. In the scenario where one of the joints is malfunctioning, the robot deviates from the calculated trajectory. So, we simplified the end-effector trajectory as a trajectory connecting three key points: the initial position ($p_i$), the drawer position ($p_d$), and the pulled-out position ($p_p$), ignoring the motion of grippers. This setup is compatible with the motion of the robot in the simulation environment. 

Given the parameters of the robot are shown in Table \mbox{\ref{tab:dh_parameters}}, and the robot is located at the coordinates of $X$ = 0, and $Y$ = 0. The end-effector goes through the initial position, the drawer position, and the pulled-out position, sequentially. The positions and Euler angles of the end-effector at each point are expressed in Table \mbox{\ref{points}}. The end-effector trajectory is solved using the iterative numerical method with a step size of 0.01 and a convergence tolerance of $1 \times 10^{-4}$.

\begin{table*}[ht]
\centering
\caption{Position of the three points constructing the end-effector trajectory: the initial position ($p_i$), the drawer position ($p_d$), and the pulled-out position ($p_p$). Euler angles ($E_x$, $E_y$, $E_z$) and translation vectors ($x$, $y$, $z$).}
    \begin{tabular}{c|cccccc}
    \toprule
     & \textbf{$E_x$ (rad)} & \textbf{$E_y$ (rad)} & \textbf{$E_z$ (rad)} & \textbf{$x$} & \textbf{$y$} & \textbf{$z$} \\
    \midrule
    \textbf{$p_i$} & $0.5 \pi$ & $0$ & $\pi$ & $0.500$ & $0$ & $0.625$ \\
    \textbf{$p_d$} & $0$ & $0$ & $0.5 \pi$ & $0.750$ & $0$ & $0.317$ \\
    \textbf{$p_p$} & $0.5 \pi$ & $0$ & $\pi$ & $0.371$ & $0$ & $0.317$ \\
    \bottomrule
    \end{tabular}
\label{points}

\end{table*}

To demonstrate the impact of a joint malfunction, we simulate the scenario where one of the joints is fixed (the third joint), i.e. it is not actuated during the task. This is equivalent to constraining the corresponding joint angle to a constant value. We then analyze the reachability and performance of the robot in completing the task of opening a drawer. The kinematic analysis is performed by solving the inverse kinematics and evaluating whether the end-effector can reach the desired position and complete the expected trajectory.

\section{Results and Discussion}
\label{sec:results}

\subsection{RL Training Results}
\begin{figure*}[ht]
    \centering
    \begin{subfigure}[b]{0.495\textwidth}
        \centering
        \includegraphics[width=\textwidth]{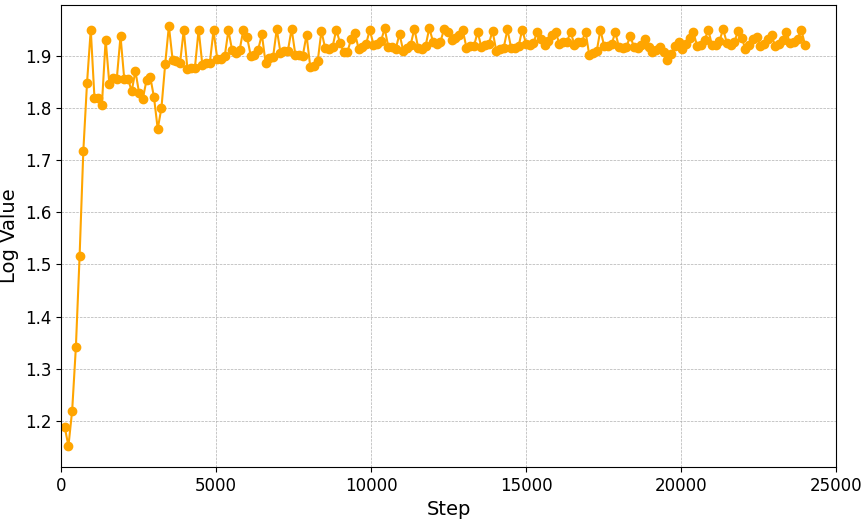}
        \caption{Distance reward.}
        \label{fig:distance}
    \end{subfigure}
    \hfill
    \begin{subfigure}[b]{0.495\textwidth}
        \centering
        \includegraphics[width=\textwidth]{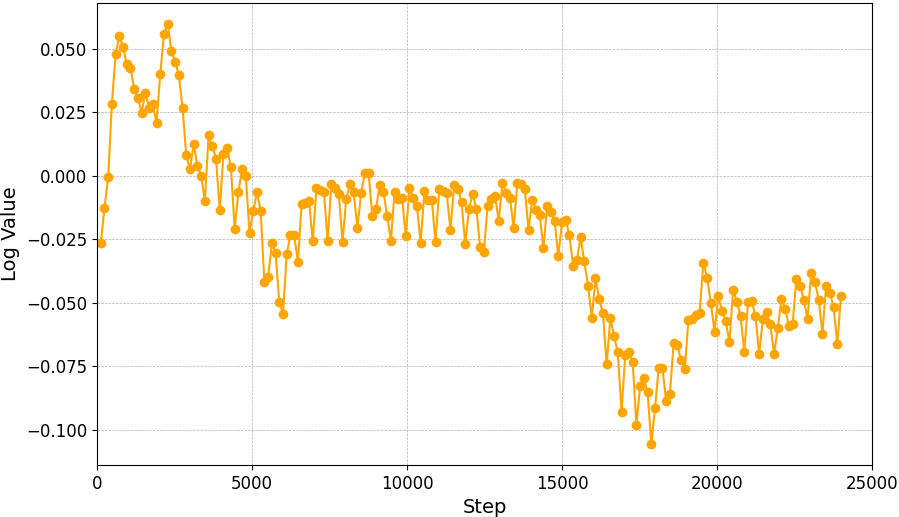}
        \caption{Rotation reward.}
        \label{fig:rot_reward}
    \end{subfigure}
    \vspace{0.5cm}
    
    \begin{subfigure}[b]{0.495\textwidth}
        \centering
        \includegraphics[width=\textwidth]{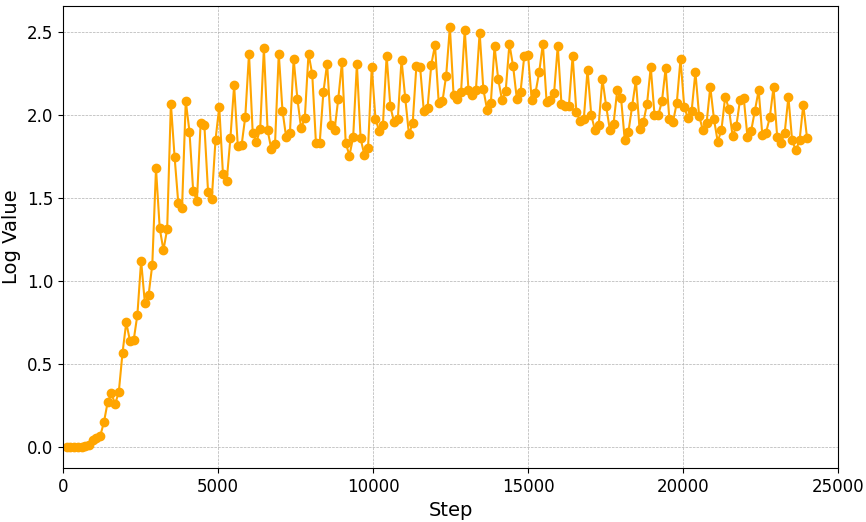}
        \caption{Opening reward.}
        \label{fig:opening}
    \end{subfigure}
    \hfill
    \begin{subfigure}[b]{0.495\textwidth}
        \centering
        \includegraphics[width=\textwidth]{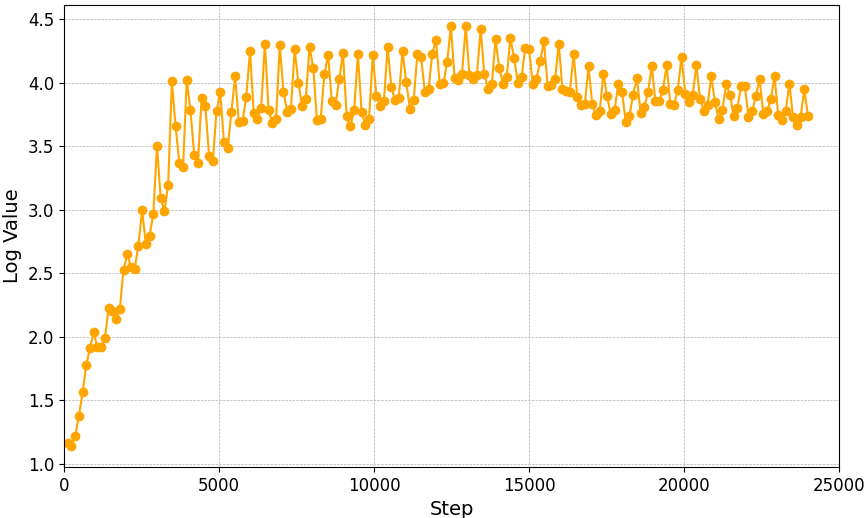}
        \caption{Total reward.}
        \label{fig:totalreward}
    \end{subfigure}    
    \caption{Reward from the training}
    \label{fig:plotsreward}
\end{figure*}

We trained the robot with 24000 episodes as described in section \mbox{\ref{sec:simulation}} using our lab computer equipped with an NVIDIA GeForce RTX 2070. The training takes around 40 minutes. The reward plot of the robot training is shown in Figure \mbox{\ref{fig:plotsreward}}, in which we show the four reward functions that we describe in section \mbox{\ref{reward_function}}.

As illustrated in Figure \ref{fig:distance}, the distance reward shows a sharp increase initially, stabilizing around the 1.8 to 1.9 log value range after approximately 2000 steps. This indicates that the robot quickly learns to minimize the distance required to open the drawer, achieving and maintaining high performance early in the training process. However, the rotation reward shows a more variable trend, with an initial increase peaking around 0.05 log value, followed by a decrease and some fluctuations, as shown in Figure \ref{fig:rot_reward}. The value eventually stabilizes around -0.075 to -0.025 log value after 20000 steps. This variability suggests that the robot had more difficulty optimizing the rotational aspect of the task compared to distance, likely due to the complexity of the rotation movements required for the task. 

Figure \ref{fig:opening} shows that the opening reward shows a consistent increase, stabilizing around a log value of 2.0 after 7500 steps. This indicates a steady improvement in the robot's ability to perform the opening action of the drawer, reflecting the effectiveness of the training in teaching the robot this aspect of the task. The total reward shown in Figure \ref{fig:totalreward}, which likely aggregates the individual rewards, demonstrates a steady increase, stabilizing around a log value of 3.5 to 4.0 after 20000 steps. This overall upward trend indicates successful training, with the robot improving its performance across all aspects of the task over time.

In addition to showing task completion rewards, we conducted experiments under different fault conditions to evaluate the performance of our RL framework under various joint malfunction scenarios. Following previous research \mbox{\cite{zhao2023learning}}, we calculate the successful task completion rate under different fault scenarios including permanently broken joints, intermittently functioning joints, joint working in the first half of testing and not working in the second half of testing, and vice versa. For the task completion rate, this metric is used to measure the effectiveness of the robot's ability to complete the drawer opening task, even in the presence of joint malfunctions. A higher completion rate indicates a more reliable and adaptable system, demonstrating the robot's capacity to overcome failures. For the time taken, we analyze the time taken to complete the task to assess the robot's efficiency. A lower time suggests more efficient performance, while an increase in time may indicate the robot has to adjust its strategy to compensate for joint malfunctions. The success rate and average completion time for each scenario are presented in Table \mbox{\ref{table:performance}}.

The results presented in Table \ref{table:performance} provide valuable insights into the robustness and efficiency of our RL framework when subjected to various joint malfunction scenarios. In the no-fault scenario, the RL algorithm demonstrates a high success rate of 98.00\% with an average completion time of 3.54 seconds. This serves as the baseline for our experiments, indicating the algorithm's effectiveness in completing the task efficiently when there are no joint malfunctions.

In addition, when a joint is permanently broken, the success rate remains high at 96.00\%, although the average completion time increases to 4.62 seconds. This increase suggests that while the algorithm compensates for the loss of functionality by adjusting its strategy, it does so at the cost of increased time.

In the case of an intermittently functioning joint, the success rate is also 96.00\%, with a slightly higher average completion time of 3.77 seconds compared to the no-fault scenario. This indicates that the algorithm can quickly adapt to intermittent faults and maintain efficiency without significant delays.

\begin{table*}[ht]
\centering
\caption{Task completion performance under different fault scenarios.}
\begin{tabular}{lcccc}
\toprule
\textbf{Fault Scenario}  & \textbf{Success Rate (\%)} & \textbf{Average Completion Time (s)} \\
\midrule
No-Fault & 98.00 & 3.54 \\
Permanently Broken Joint & 96.00 & 4.62\\
Intermittently Functioning Joint & 96.00 & 3.77 \\
Joint Works First Half & 96.00 & 4.11 \\
Joint Works Second Half & 82.00& 8.02\\
\bottomrule
\end{tabular}

\label{table:performance}
\end{table*}

When the joint functions only during the first half of the task, the success rate remains at 96.00\%, with an average completion time of 4.11 seconds. This result implies that the RL algorithm effectively utilizes the functional period of the joint to complete the task, demonstrating its ability to optimize performance under partial functionality. 

\begin{figure*}[ht]
    \centering    
    \begin{subfigure}[b]{0.495\textwidth}
        \centering
        \includegraphics[width=\textwidth]{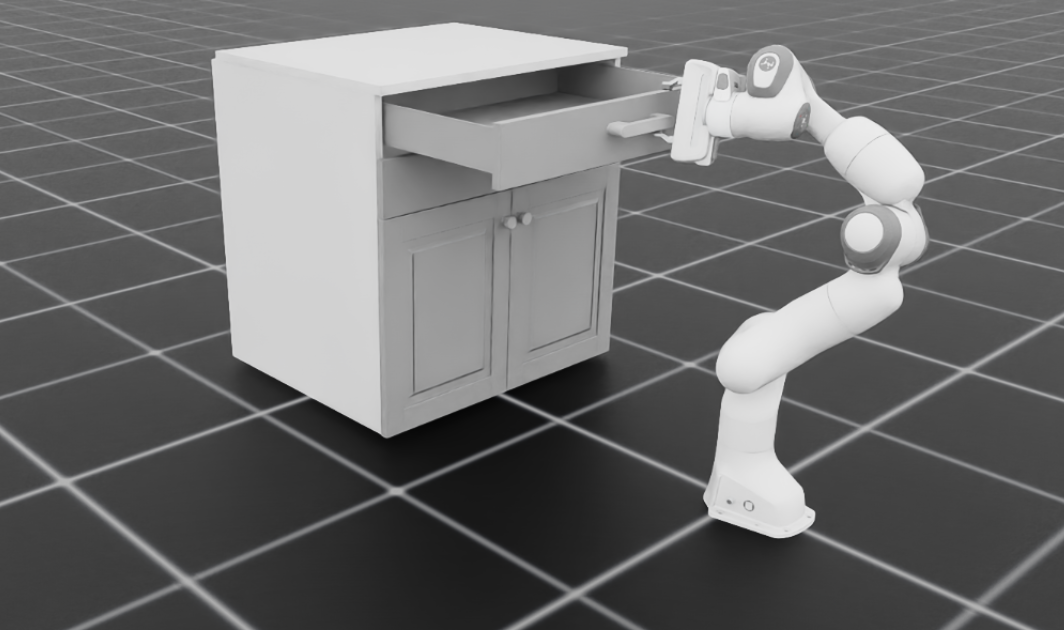}
        \caption{The robot successfully opens the cabinet.}
        \label{fig:success_case}
    \end{subfigure}
    \hfill
    \begin{subfigure}[b]{0.495\textwidth}
        \centering
        \includegraphics[width=\textwidth]{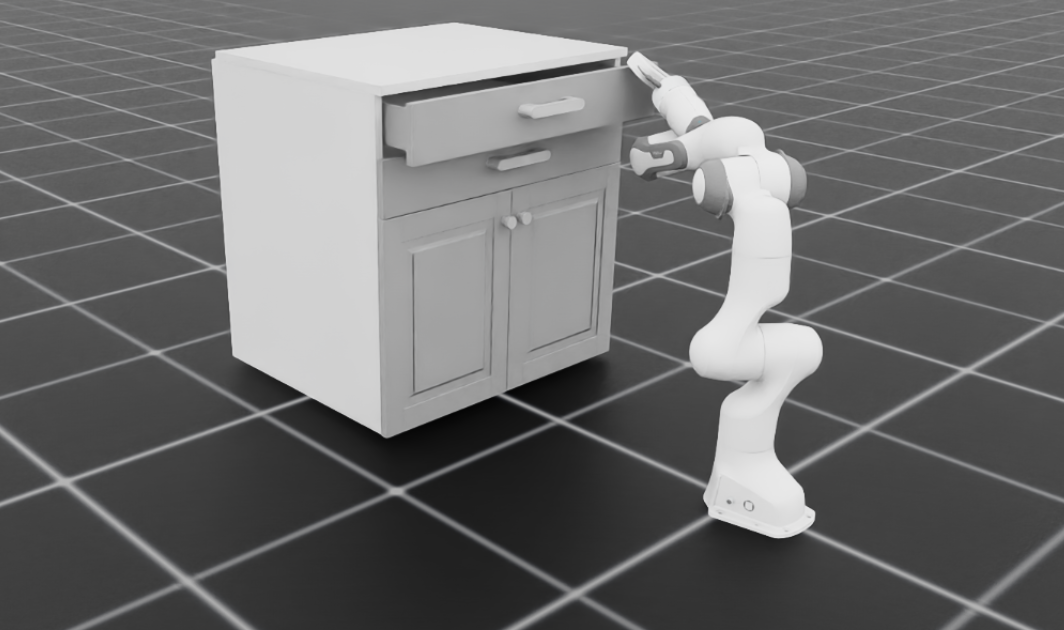}
        \caption{The robot fails to open the cabinet.}
        \label{fig:failcase}
    \end{subfigure}
    \caption{Example of successful task completion and failed task completion.}
    \label{fig:Example}
\end{figure*}

However, the scenario where the joint works only in the second half presents the most significant challenge. The success rate drops to 82.00\%, and the average completion time increases substantially to 8.02 seconds. The reduced success rate and longer completion time indicate that the algorithm struggles more when the joint only functions in the latter part of the task. This difficulty likely arises from the initial lack of joint functionality in the first half, requiring the algorithm to employ more complex strategies to compensate, leading to longer task completion times and lower success rates. Figure \ref{fig:Example} shows two cases, one the robot successfully opens the cabinet, and the other the robot fails to open the cabinet.

Generally, our framework demonstrates strong adaptability and robustness across different fault scenarios, maintaining high success rates in most cases. The completion time varies depending on the type and timing of the joint malfunction, with intermittent faults and partial functionality scenarios resulting in moderate increases in time. The scenario where the joint works only in the second half poses the greatest challenge, highlighting an area for potential improvement in the algorithm's adaptability to late-functioning components. However, these results underscore the effectiveness of the RL algorithm in handling joint malfunctions, ensuring reliable task completion even under adverse conditions.

\subsection{Kinematics Results}

\begin{figure*}[ht]
    \centering    
    \begin{subfigure}[b]{0.495\textwidth}
        \centering
        \includegraphics[width=\textwidth]{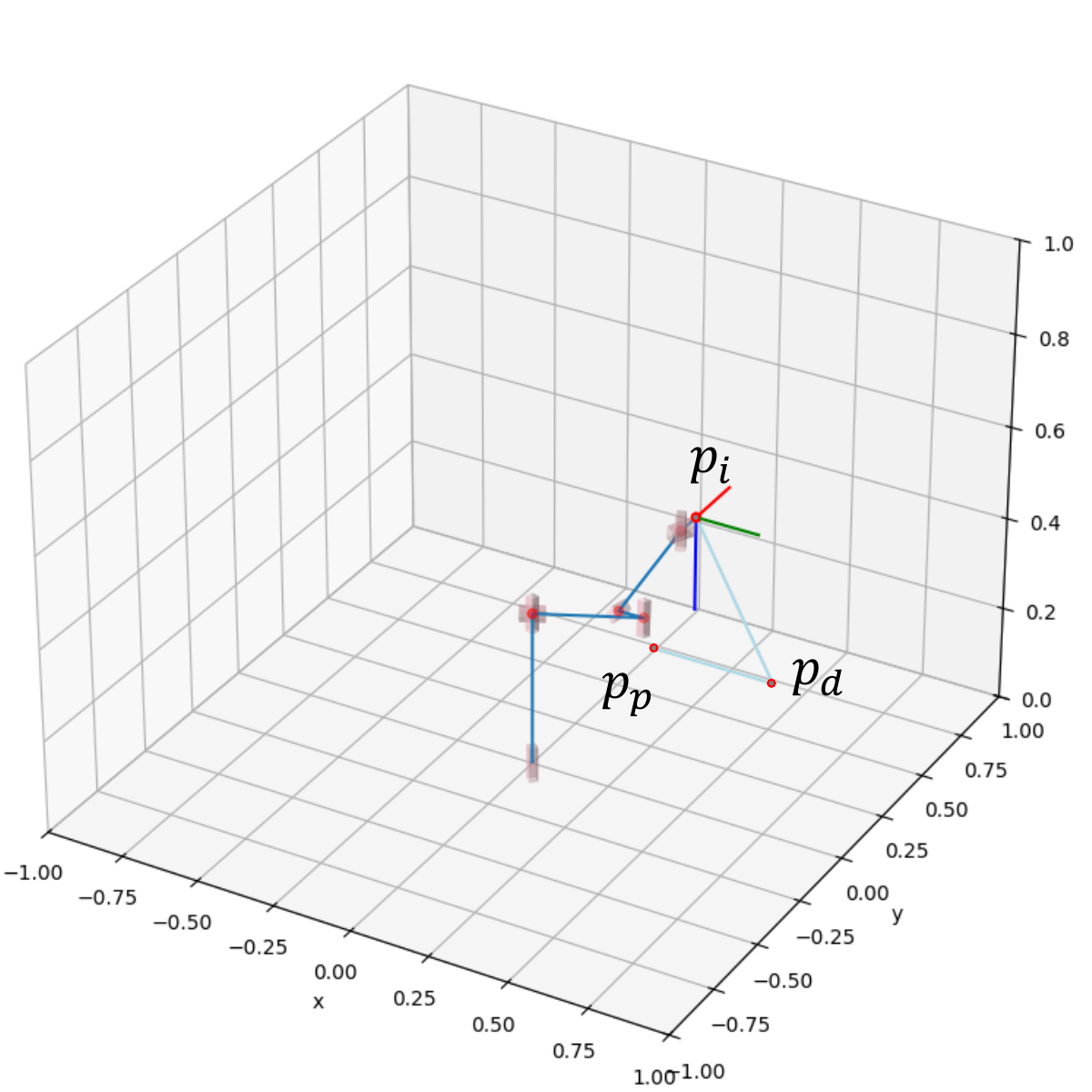}
        \caption{The end-effector follows the desired trajectory when all joints of the robot work properly.}
        \label{fig:kinematic_working}
    \end{subfigure}
    \hfill
    \begin{subfigure}[b]{0.495\textwidth}
        \centering
        \includegraphics[width=\textwidth]{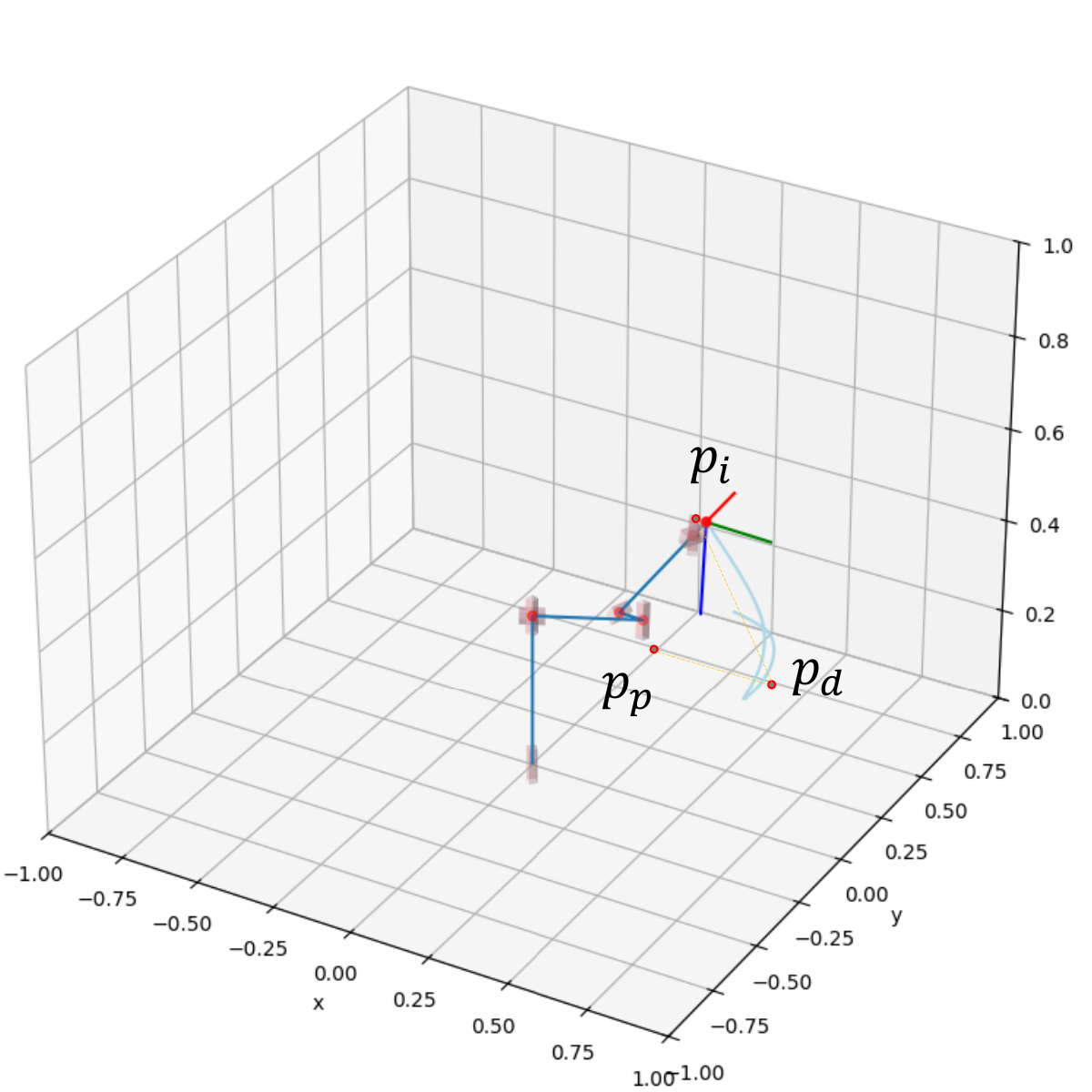}
        \caption{The end-effector fails to follow the expected trajectory when one of the joints is broken.}
        \label{fig:kinematic_failure3}
    \end{subfigure}
    \caption{Comparison of the trajectories of the end-effector when one of the joints of the robot is broken and when all the joints work properly. Typically, when the robot operates properly, it must follow the desired trajectory (orange lines) constructed by three joints: the initial position ($p_i$), the drawer position ($p_d$), and the pulled-out position ($p_p$).}
    \label{fig:Kinematics}
\end{figure*}

In addition to our RL-based control method, we also demonstrate that with a broken joint, the robot's end-effector is unable to follow the programmed trajectory. The details of the experimental setup are discussed in section \mbox{\ref{kinematic_setup}}, and intuitive results are presented in Figure \mbox{\ref{fig:Kinematics}}. Figure \mbox{\ref{fig:kinematic_working}} shows that the robot follows the trajectory from the initial position to the drawer position and opens it, which goes through the three points $p_i$, $p_d$, and $p_p$. However, as shown in Figure \mbox{\ref{fig:kinematic_failure3}}, when one joint is fixed, typically joint 3, the robot is unable to reach the desired end-effector trajectory, failing to open the drawer.

The results demonstrate the necessity of all joints being operational properly for completing a task. The kinematic analysis using DH parameters and inverse kinematics highlights the critical role of each joint in the Franka robot's ability to perform complex tasks. The demonstrated impact of a joint malfunction underscores the importance of developing robust control algorithms that can compensate for such failures \mbox{\cite{survey_ftc}}. By developing a DRL algorithm based on our proposed framework, we successfully demonstrate that joint failures can be effectively handled in robotic manipulators.

\section{Conclusion and Future Work}
\label{sec.6conclusion}

In this study, we addressed the challenge of enabling a robotic manipulator to complete tasks despite joint malfunctions by developing a reinforcement learning framework. Our experimental platform, the Franka robot with 7 degrees of freedom, was used to test the framework under various joint failure conditions, including permanently broken and intermittently functioning joints. By framing the problem as a partially observable Markov decision process, we successfully trained the robot to adapt to these varying conditions.

In the downstream task, we tested the RL algorithm in both seen and unseen cases including the joint working in the first half of the time and vice versa. To evaluate the robot's performance, we calculate the task completion rate and the time taken. Our results demonstrated that the RL-trained robot was able to complete the drawer opening task even with joint failures, achieving a high success rate with an average rate of 93.6\% and an average operation time of 3.8s. The results highlight the adaptability and resilience of our approach. Furthermore, the RL algorithm showed strong performance in both seen and unseen failure scenarios, indicating its potential for real-world applications where unexpected hardware failures are common. Compared to the kinematic-based control, if one of the joints is non-functional, the robot's end-effector is unable to follow the desired trajectory, leading to task failure.

This study underscores the potential of RL to enhance the reliability of robotic systems, making them better suited for unpredictable environments. By integrating RL with advanced simulation environments like Isaac Lab, we have shown that it is possible to create robust and adaptable robotic solutions that can maintain functionality despite hardware malfunctions. While our approach shows promise for practical applications, there are still limitations to deploying RL-based systems in real-world manufacturing environments. The computational complexity and training time required for RL can present challenges, especially for completing complex tasks.

In the future, we aim to transition our framework from simulation to real-world applications. Furthermore, future efforts could extend this approach to other types of robotic tasks such as pick-and-place objects. The integration of more sophisticated RL algorithms and neural networks such as Transformer-based models or CNN-based models could be considered to further improve performance and adaptability.

\balance
{\small
\bibliographystyle{ieee_fullname}
\bibliography{egbib}
}

\end{document}